# C2W-Tune: Cavity-to-Wall Transfer Learning for Thin Atrial Wall Segmentation in 3D LGE-MRI

Yusri Al-Sanaani[1], Rebecca Thornhill[1,2], and Sreeraman Rajan[1]

[1] Systems and Computer Engineering, Carleton University, Ottawa, Canada
[2] Department of Radiology, University of Ottawa, Ottawa, Canada
YusriAlSanaani@cmail.carleton.ca

**Abstract.** Accurate segmentation of the left atrial (LA) wall in 3D late gadolinium-enhanced MRI (LGE-MRI) is essential for wall thickness mapping and fibrosis quantification, yet it remains challenging due to the wall's thin geometry, complex anatomy, and low contrast. We propose C2W-Tune, a two-stage cavity-to-wall transfer framework that leverages a high-accuracy LA cavity model as an anatomical prior to improve thin-wall delineation. Using a 3D U-Net with a ResNeXt encoder and instance normalization, Stage 1 pre-trains the network to segment the LA cavity, learning robust atrial representations. Stage 2 transfers these weights and adapts the network to LA wall segmentation using a progressive layer-unfreezing schedule to preserve cavity features while enabling wall-specific refinement. On the 2018 LA Segmentation Challenge dataset, C2W-Tune outperformed an architecture-matched baseline trained from scratch. The wall Dice score increased from 0.623 to 0.814, surface Dice at 1 mm increased from 0.553 to 0.731, 95th-percentile Hausdorff distance (HD95) decreased from 2.95 mm to 2.55 mm, and average symmetric surface distance (ASSD) decreased from 0.71 mm to 0.63 mm. Under reduced supervision using 70 training volumes sampled from the same training set, C2W-Tune achieved a Dice of 0.78, remaining competitive with recent multi-class bi-atrial benchmarks (typically 0.6–0.7). These results show that anatomically grounded task transfer with controlled fine-tuning improves accuracy for thin LA wall segmentation in 3D LGE-MRI.

## 1 Introduction

Late gadolinium-enhanced MRI (LGE-MRI) is widely used to assess and quantify scarring within the left atrium (LA). The spatial distribution and severity of these fibrotic regions may offer key insights into the underlying mechanisms and clinical progression of atrial fibrillation (AF). Consequently, robust computational methods for processing and analyzing LA LGE-MRI data are important for supporting image-guided diagnosis and guiding personalized treatment strategies for patients with AF [1]. For example, LGE-MRI can enable measurement of LA wall thickness, which predicts ablation outcome and safety, while delineation of the endocardial (inner) and epicardial (outer) surfaces enables quantification of scar transmurality (full-thickness injury), which correlates with AF recurrence [1]. However, automatically segmenting the LA wall from 3D LGE-MRI is extremely challenging. The LA wall is very thin and has a complex geometry with multiple discontinuities (e.g., pulmonary veins (PV) and the mitral valve (MV)), making it difficult to distinguish it from surrounding structures. LGE-MRI images often suffer from low contrast and heterogeneous enhancement; the atrial wall and nearby tissues can exhibit similar intensity, complicating boundary identification [1–4]. These factors lead to severe class imbalance and segmentation ambiguity because the LA wall is very thin, and even expert, fellowship-trained cardiothoracic radiologists struggle to provide reproducible manual delineations [5].

As a result, most prior work has focused on the more tractable task of segmenting the LA cavity. For instance, the 2018 LA Segmentation Challenge benchmark demonstrated that a two-step pipeline (localization followed by refinement) achieved a Dice score of 93.2% and a mean surface distance of approximately 0.7 mm on the test set [2]. Puybareau et al. achieved rapid left atrial cavity segmentation in 3D LGE-MRI by fine-tuning an ImageNet-pretrained VGG-16 fully convolutional network on the 2018 LA Segmentation Challenge dataset using a pseudo-3D slice-stacking input [6]. These works indicate that with sufficient training data and appropriate architecture, the LA cavity (blood pool) can be segmented with high accuracy from LGE-MRI.

While deep learning has advanced cavity segmentation [2] and joint cavity-scar frameworks [7], automating thin atrial wall segmentation remains a major bottleneck. The recent multi-class bi-atrial segmentation challenge (MBAS) further highlights this performance gap. It evaluated a wide spectrum of architectures, including 3D nnU-Nets with residual encoders such as [8, 9], Mamba-style state-space models such as [9, 10], transformer-augmented networks such as [9, 11], and various multi-stage pipelines such as [12, 13]. Across these methods, the left and right cavities consistently achieve Dice scores above 0.90, whereas the bi-atrial (left and right) wall typically remains in the ~0.6–0.7 Dice range. Although most MBAS approaches formulate the problem as the joint prediction of the LA cavity, right atrium (RA) cavity, and bi-atrial wall within a single model, this shared-parameter training induces multi-task-like interactions among outputs. In practice, atrial wall delineation remains the most challenging target, with consistently lower Dice than the cavity classes across challenge reports, suggesting that simultaneous optimization may be biased toward the easier cavity classes.

Several methods implicitly address this imbalance through coarse-to-fine pipelines that refine wall predictions after an initial joint segmentation, such as [12] and [13]. Recent work, such as [14], similarly adopts a two-stage ResNeXt-based architecture using a coarse-to-fine spatial cropping strategy for bi-atrial segmentation, but the left and right walls are segmented separately. Conventional deep learning workflows in medical imaging, such as [6], often rely on transfer learning, in which a model is pre-trained on a large, unrelated natural-image dataset (e.g., ImageNet) and then fine-tuned on the target medical task [15]. While this strategy can effectively initialize low-level feature detectors (e.g., edges and blobs), it does not necessarily encode the domain-specific anatomical priors needed for robust medical image analysis.

In this work, we propose C2W-Tune (Cavity-to-Wall Tune), a two-stage 3D segmentation approach that leverages the strong performance of cavity segmentation by initializing wall segmentation from a high-quality cavity model via task-specific anatomical transfer learning. Instead of learning wall segmentation from scratch, C2W-Tune uses the pre-trained LA cavity segmentation model as an anatomical prior, which is subsequently fine-tuned for the thin-wall segmentation task by replacing the final segmentation head and progressively adapting the transferred network for wall segmentation. By transferring the learned encoder features from the cavity task, we prime the network to recognize the LA region, while the progressive fine-tuning avoids catastrophic forgetting of these features.

## 2 Methodology

Our C2W-Tune approach consists of two stages. First, we use a 3D U-Net to produce a coarse segmentation that localizes the atrial region. We then compute the center of mass of the coarse atrial mask and extract a fixed patch around it to define the region of interest (ROI). Stage 1 trains a second 3D U-Net to segment the LA cavity from these ROIs. This model provides a robust encoder pre-trained on the anatomical region of interest. Stage 2 fine-tunes this model to segment the LA wall by introducing a new segmentation head and employing a progressive layer-unfreezing schedule. **Fig. 1** illustrates the pipeline.

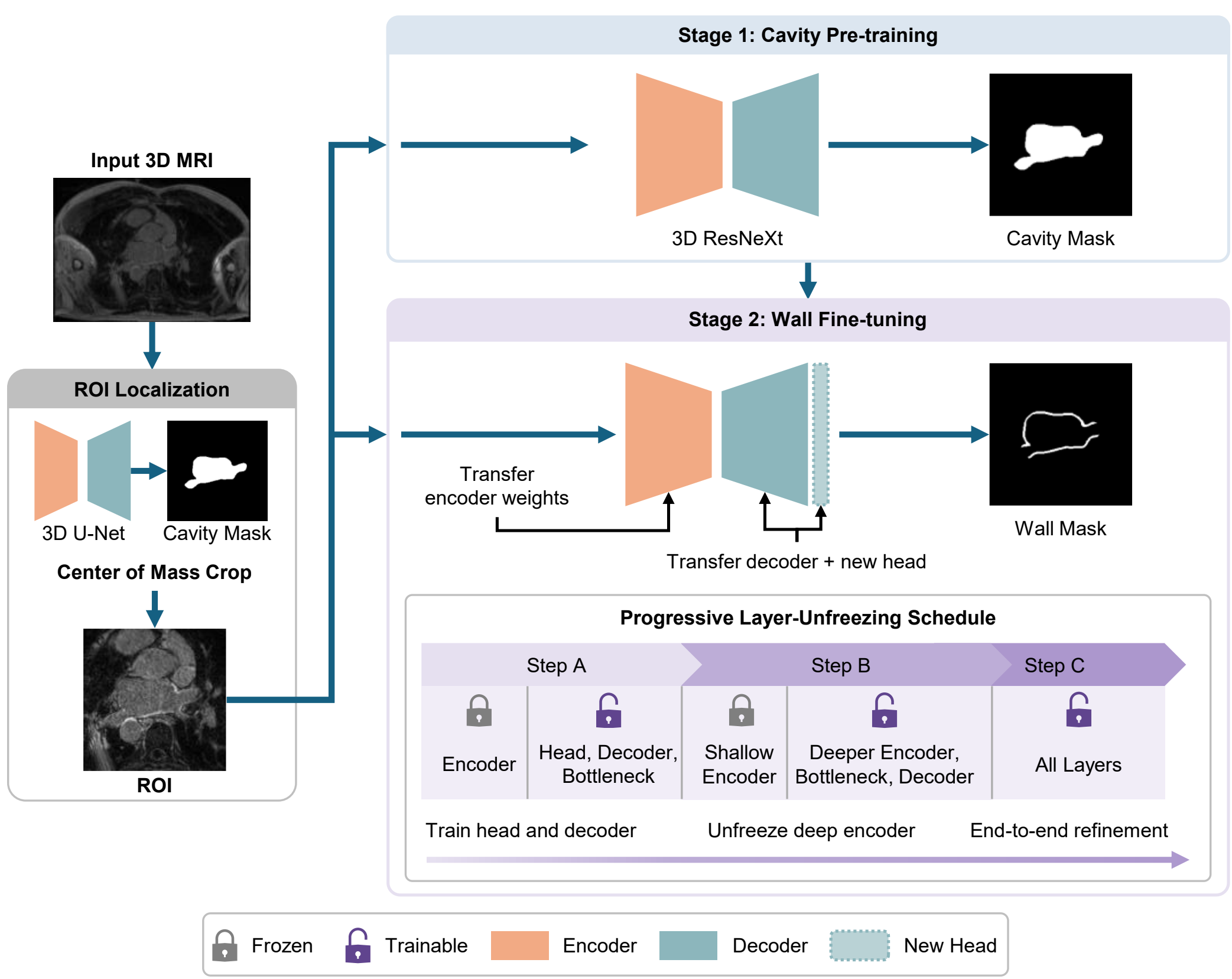

**Fig. 1.** C2W-Tune pipeline: atrial ROI localization, cavity pre-training, and wall fine-tuning.

### 2.1 Dataset

Let $\mathcal{D} = \{(X_i, Y_i^{cav}, Y_i^{wall})\}_{i=1}^{N}$ be the dataset of N 3D LGE-MRI volumes. $X_i \in R^{D\times H\times W}$ is the input LGE-MRI volume. $Y_i^{cav} \in \{0,1\}^{D\times H\times W}$ is the binary ground truth for the LA cavity. $Y_i^{wall} \in \{0,1\}^{D\times H\times W}$ is the binary ground truth for the LA wall. We utilize the 2018 LA Segmentation Challenge dataset [2]. The dataset consists of 154 LGE-MRI 3D volumes of patients with AF, acquired at an isotropic voxel spacing of approximately 0.625 mm. Each volume has an expert manual segmentation of the LA cavity and LA wall as ground truth masks.

To reduce background interference and mitigate class imbalance, we first apply a coarse 3D U-Net to each input volume $X_i$ to obtain a preliminary cavity prediction $\hat{Y}_i^{cav}$. We then compute the center of mass of $\hat{Y}_i^{cav}$ and crop $X_i$ to a fixed ROI patch of size 256 × 256 × 44 voxels (and cropping $Y_i^{cav}$and $Y_i^{wall}$ identically during training). The ROI size is chosen to preserve sufficient spatial context around the atrial wall, consistent with findings of the 2018 LA Segmentation Challenge benchmark [2], which reported performance degradation for ROIs smaller than 240×160. Therefore, this ROI restricts subsequent cavity/wall prediction to an anatomically relevant subvolume, reducing the effective search space and encouraging the model to focus on left atrial structures rather than surrounding tissue.

### 2.2 Network Architecture

We employ a 3D U-Net with a ResNeXt-based encoder [16] to segment thin atrial structures. In this design, the grouped convolutions aggregate multiple feature transformations efficiently, improving representation quality without parameter explosion. To stabilize training with small batch sizes, instance normalization is used throughout. The encoder consists of seven stages, and each stage uses 3×3×3 convolution kernels. The network initializes with 32 features in the first layer. The number of features doubles progressively in the early stages (64, 128, and 256). The final three deep stages of the encoder maintain a consistent channel depth of 512 features. This configuration promotes effective capture of high-level anatomical representations of the atrial anatomy while preserving the fine-grained spatial details critical for delineating the cavity and the thin atrial wall.

### 2.3 Stage 1: Cavity Prior Learning

In the first stage, we pre-train the network on the source task to learn a robust cavity prior, focusing on the global atrial geometry and its high-contrast boundaries. Given the input volume $X_i$, the network is trained to predict the corresponding atrial cavity mask $\hat{Y}_i^{cav}$. This stage is designed to capture anatomical structure and reliable boundary cues. Training is performed for 1000 epochs with early stopping. Upon convergence, the trained encoder–decoder parameters, denoted by $\theta_{\mathrm{cav}}$, are retained.

### 2.4 Stage 2: Fine-Tuning for LA Wall Segmentation

In the second stage, we initialize the wall-segmentation network ($\theta_{wall}$) by transferring anatomical representations learned during cavity pre-training ($\theta_{cav}$). This transfer provides a strong starting point for wall prediction by preserving the learned global atrial geometry and contextual features. We then fine-tune the model for atrial wall segmentation using the progressive layer-unfreezing schedule described below.

### 2.5 Progressive Unfreezing Schedule

To mitigate catastrophic forgetting during the transfer from cavity to wall segmentation, we adopt a three-step progressive layer-unfreezing schedule strategy. This strategy leverages layer-wise fine-tuning [17] to incrementally adapt the network, thereby preserving learned feature representations and reducing the risk of losing global atrial geometry during optimization.

**Step A (Head/Decoder Remapping; epochs 0–60).** We freeze the entire encoder and train the wall segmentation head, decoder, and bottleneck, allowing the decoder to remap transferred atrial features from cavity to wall prediction. In this stage, the encoder serves as a fixed feature extractor, while the decoder is adapted to reinterpret the transferred atrial features for wall prediction rather than cavity prediction.

**Step B (Deep Fine-tuning; epochs 61–180).** We keep the shallow encoder layers frozen and unfreeze the deeper encoder stages, together with the bottleneck and decoder. This allows higher-level semantic representations to adapt to wall-specific appearance and geometry, while maintaining stable low-level edge and intensity features learned during pre-training.

**Step C (Full Optimization; epochs 181–Max 1000).** Finally, all layers are unfrozen for end-to-end refinement. Early stopping is applied during Step C based on validation performance, and the best-performing checkpoint is retained.

The epoch boundaries (e.g., 60 and 180) are validation-selected hyperparameters rather than fixed constants.

### 2.6 Training Setup

The LGE-MRI images were normalized using z-score normalization. The dataset was split into 30 independent test cases (held-out for final evaluation) and 124 development cases. To optimize hyperparameters and assess model stability without data leakage, we performed five-fold cross-validation on the 124 development cases. All splits were performed at the patient level. Following hyperparameter optimization via cross-validation, the final model was retrained once using a fixed 100/24 train/validation split from the 124 development cases and evaluated on the held-out test set. Data augmentation was applied on-the-fly to improve generalization. We used geometric augmentations (random elastic deformations to simulate anatomical variability, and random rotations and axis-constrained flips to reflect plausible pose and orientation differences) and intensity augmentations (intensity scaling and histogram matching to mitigate scanner- and protocol-dependent intensity variability in LGE-MRI [18]). These augmentations help reduce overfitting given the dataset size and address potential domain shifts. Training used the AdamW optimizer with $\beta_1 = 0.9$, $\beta_2 = 0.999$, and a weight decay of $1 \times 10^{-4}$. We used DiceFocal loss [19] to mitigate class imbalance for both stages, and a batch size of 4. For Stage 1, we utilized a cosine annealing schedule with a 50-epoch linear warmup to stabilize the grouped convolutions, decaying the learning rate from $1 \times 10^{-3}$ to $1 \times 10^{-6}$. For Stage 2, given the constraint of small batch sizes, we employed a stage-wise cosine annealing scheduler with linear warmup. Consistent with our progressive unfreezing strategy, the maximum learning rate was decayed by a factor of 10 at each restart ($1 \times 10^{-3} \rightarrow 1 \times 10^{-4} \rightarrow 1 \times 10^{-5}$) to avoid disrupting the transferred spatial representations during the transition from cavity to wall segmentation. Hyperparameters were selected empirically based on validation Dice (five-fold cross-validation).

### 2.7 Evaluation Metrics

We report the Dice Similarity Coefficient to measure global volumetric overlap between the predicted and reference masks. To emphasize boundary accuracy, we additionally compute surface Dice at 1 mm, defined as the fraction of surface points on the predicted and ground-truth masks that lie within 1 mm of each other; this is particularly informative for thin-wall segmentation [20]. Finally, we report HD95 (95th-percentile Hausdorff distance) to capture near-worst-case boundary deviations while reducing sensitivity to extreme outliers, together with average symmetric surface distance (ASSD) to summarize the average symmetric surface error.

## 3 Results and Discussion

To quantify the difficulty of LA wall segmentation, we trained an architecture-matched 3D U-Net baseline from scratch using the same loss, data split, and training protocol, but without cavity pre-training. Stage 1 produced a strong cavity model, achieving a Dice score of 0.921 on the cavity test set, comparable to published results [2, 14]. These weights were then transferred for wall segmentation in Stage 2. As shown in **Table 1**, the baseline had difficulty delineating the thin atrial wall, yielding a Dice score of 0.623 and often producing fragmented boundaries. In contrast, C2W-Tune effectively leveraged the anatomical features learned during Stage 1 to guide the wall segmentation. This resulted in a substantial performance gain: wall Dice increased to 0.814 (paired two-sided Wilcoxon signed-rank, $p = 0.001$ vs. baseline model), surface Dice increased from 0.55 to 0.73, HD95 decreased from 2.95 mm to 2.55 mm, and ASSD decreased from 0.71 mm to 0.63 mm. These metrics indicate that C2W-Tune improves overlap and boundary agreement, generating continuous, realistic surfaces compared to the fragmented predictions of the baseline model. **Fig. 2** shows representative axial slices from challenging cases, including thin low-contrast walls and complex geometry near the pulmonary vein ostia and appendage. In these examples, the baseline often exhibits fragmentation or missing segments, whereas C2W-Tune better preserves boundary continuity.

**Table 1.** LA wall segmentation performance on the held-out test set (mean ± standard deviation across test cases).

| Method | Wall Dice ↑ | Surface Dice ↑ | HD95 (mm) ↓ | ASSD (mm) ↓ |
|---|---|---|---|---|
| 3D U-Net (baseline) | 0.623 ± 0.042 | 0.553 ± 0.061 | 2.95 ± 0.88 | 0.71 ± 0.24 |
| C2W-Tune | **0.814 ± 0.028** | **0.731 ± 0.045** | **2.55 ± 0.55** | **0.63 ± 0.17** |

These results indicate that using a well-resolved anatomical target (the LA cavity) to guide the more difficult wall segmentation helps address the challenges of thin LA wall delineation. Stage 1 provides a strong anatomical prior by identifying the atrial region and approximating the endocardial boundary. The progressive layer-unfreezing fine-tuning strategy then preserves this anatomical context while enabling adaptation to better capture the epicardial boundary (outer wall border).

A direct numerical comparison with prior studies is contextual due to differences in task definitions and data partitions. As detailed in **Table 2**, the majority of recent benchmarks [8, 9, 11–13] target a combined bi-atrial wall class (LA+RA) within a multi-class framework, whereas our method focuses specifically on the clinically relevant LA wall. Furthermore, many of these approaches were evaluated using restricted training sets (typically $N \approx 70$). To ensure a fair comparison regarding supervision size, we performed a controlled ablation by retraining C2W-Tune using 70 randomly selected volumes from the 100-volume training pool, while keeping the validation and test sets fixed. Results are reported as mean ± standard deviation across multiple 70-volume subsets; prior methods are reported as published because variance was unavailable (**Table 2**). Under matched supervision, C2W-Tune (N=70) achieved 0.78 Dice and 3.15 mm HD95, compared with published Dice values of 0.62–0.71 for recent multi-class bi-atrial methods and 0.617 for LA wall in [14].

These results suggest that C2W-Tune benefits from leveraging the cavity prior through progressive fine-tuning, rather than from training-set size alone. Consequently, our LA wall performance remains competitive in reduced supervision settings and compares favorably to reported ranges, while our primary comparison against the scratch baseline remains a fully controlled evaluation (**Table 1**).

**Table 2.** Contextual comparison with state-of-the-art bi-atrial segmentation methods.

| Method | Target Label | Train N | Wall Dice ↑ | HD95 (mm) ↓ |
| --- | --- | --- | --- | --- |
| 3D nnU-Net Ensemble [8] | bi-atrial | 70 | 0.672 | 4.24 |
| 3D MambaBot [9] | bi-atrial | 70 | 0.712 | 3.21 |
| 3D nnU-Net [11] | bi-atrial | 70 | 0.670 | 4.53 |
| 3D nnU-Net [12] | bi-atrial | 70 | 0.626 | 5.21 |
| SegBAW-Net [13] | bi-atrial | 70 | 0.620 | 6.32 |
| 3D U-Net [14] | LA wall | 80 | 0.617 | 3.42 |
| C2W-Tune (ours) | LA wall | **70** | **0.78 ± 0.02** | **3.15 ± 0.21** |

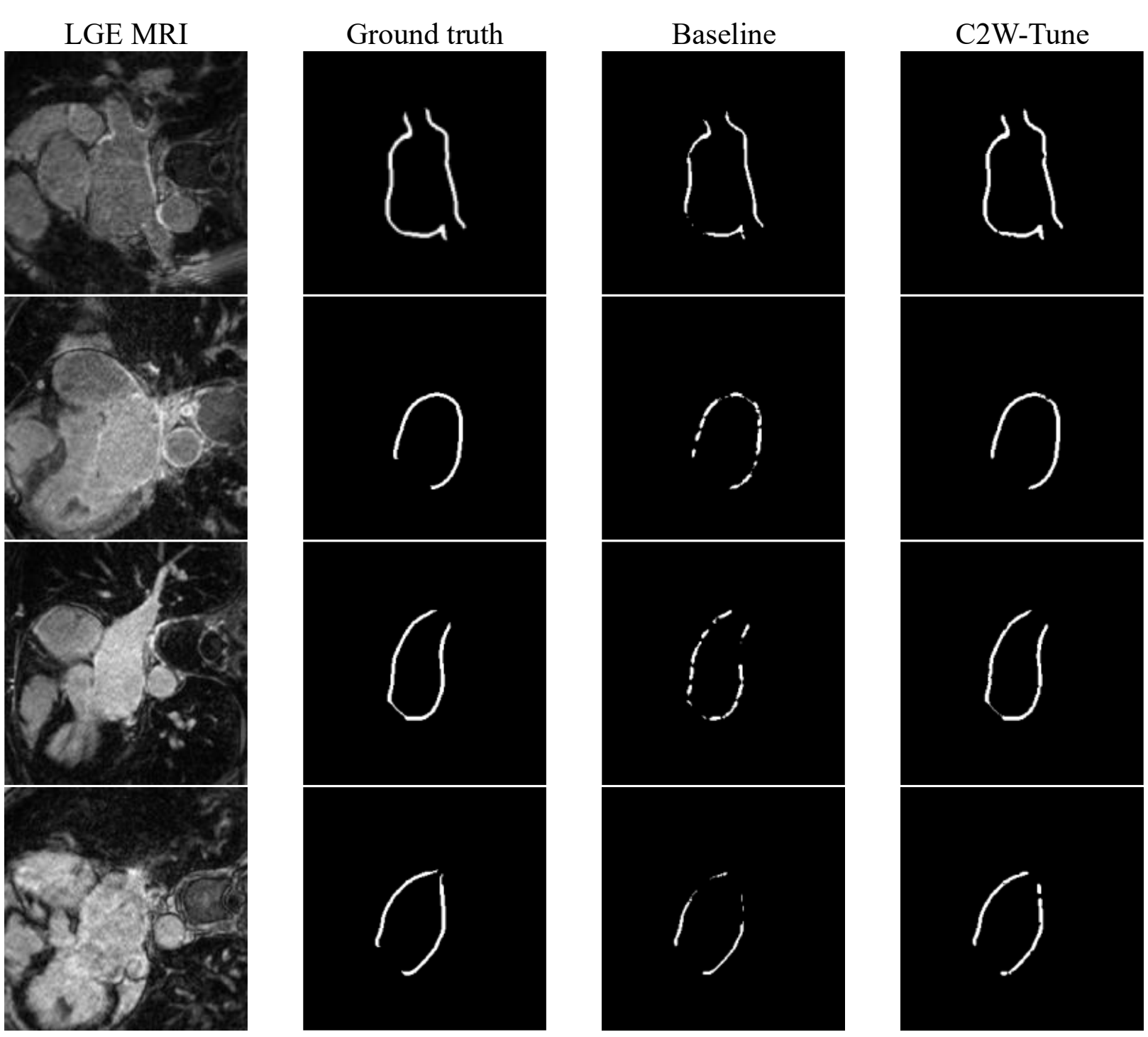

**Fig. 2.** Representative axial slices comparing ground truth, baseline, and C2W-Tune.

**Limitations and Future Work.** Evaluation used a single public dataset with a fixed split, so generalization to external cohorts, scanners, and protocols remains to be assessed. In addition, C2W-Tune was evaluated only for LA wall segmentation, and its extension to RA wall or combined bi-atrial wall targets requires further validation because of anatomical and appearance differences. Although the progressive layer-unfreezing strategy improved boundary quality, residual errors may still occur in challenging regions such as the pulmonary vein ostia and LA appendage, where boundaries are ambiguous and inter-observer variability is high. Future work will focus on external multi-center validation, domain-shift analysis, extension to bi-atrial and multi-class settings, and explicit boundary-aware refinements, including inner/outer wall-surface supervision and boundary-sensitive objectives.

## 4 Conclusion

We presented C2W-Tune, a two-stage cavity-to-wall transfer framework for thin LA wall segmentation in 3D LGE-MRI. The method first learns robust atrial representations via LA cavity pre-training and then adapts the same architecture to wall segmentation using a progressive layer-unfreezing curriculum, which preserves informative endocardial features while enabling wall-specific refinement. On held-out test set, C2W-Tune achieved higher overlap and improved boundary accuracy than an architecture-matched baseline trained from scratch and remained competitive under reduced supervision settings. Furthermore, this work suggests that hierarchical task transfer, in which an easier anatomical target guides a more difficult one, is a promising strategy for addressing extreme thin-structure imbalance in LGE-MRI. These improvements in surface fidelity and topological continuity may enhance patient-specific wall analysis and atrial remodeling assessment, supporting more reliable fibrosis/thickness mapping, more precise image-guided ablation planning, and AF recurrence risk stratification.

**Acknowledgments.** This research was enabled in part by the Natural Sciences and Engineering Research Council of Canada (NSERC) Discovery Grant, and by computational resources provided by the Digital Research Alliance of Canada (https://alliancecan.ca).